\documentclass{article}

     \PassOptionsToPackage{round}{natbib}

 \usepackage[final]{neurips_2019}

\usepackage{latexsym}
\usepackage[]{algorithm2e}
\usepackage{url}
\usepackage{booktabs}
\usepackage{amsmath}
\usepackage{graphicx}
\usepackage{wrapfig}
\usepackage{caption,subcaption}
\usepackage{todonotes}
\usepackage{forest}
\usepackage{hyperref}
\usepackage{soul}
    \usepackage{fixltx2e}
    \MakeRobust{\overrightarrow}
\newcommand{\cell}[0]{\mathrm{cell}}
\newcommand{\cp}[0]{\overrightarrow{\pi}}
\newcommand{\rcp}[0]{\overleftarrow{\pi}}
\newcommand{\vg}[0]{v}
\newcommand{\hg}[0]{h}
\newcommand{\cg}[0]{c}

\forestset{
    nice empty nodes/.style={
        for tree={
            s sep=0.1em, 
            l sep=0.33em,
            inner ysep=0.4em, 
            inner xsep=0.05em,
            l=0,
            calign=midpoint,
            fit=tight,
            where n children=0{
               tier=word,
               minimum height=1.25em,
            }{},
            where n children=2{
               l-=1em,
            }{},
            parent anchor=south,
            child anchor=north,
            delay={if content={}{
                    inner sep=0pt,
                    edge path={\noexpand\path [\forestoption{edge}] 
                    			(!u.parent anchor) 
                               -- (.south)\forestoption{edge label};}
                }{}}
        },
    },
}

\title{Ordered Memory}

\author{
  Yikang Shen\thanks{\texttt{yi-kang.shen@umontreal.ca, tanjings@mila.quebec, arian.hosseini9@gmail.com}.} \\
  Mila/Universit\'e de Montr\'eal \\ 
  and Microsoft Research \\
  Montr\'eal, Canada \\
  \And
  Shawn Tan$^{*}$\\
  Mila/Universit\'e de Montr\'eal \\
  Montr\'eal, Canada \\
  \And
  Arian Hosseini$^{*}$\\
  Mila/Universit\'e de Montr\'eal \\
  and Microsoft Research \\
  Montr\'eal, Canada \\
  \And
  Zhouhan Lin\\
  Mila/Universit\'e de Montr\'eal \\
  Montr\'eal, Canada \\
  \And
  Alessandro Sordoni \\
  Microsoft Research \\
  Montr\'eal, Canada \\
  \And
  Aaron Courville \\
  Mila/Universit\'e de Montr\'eal \\
  Montr\'eal, Canada \\
}

\begin{document}

\maketitle

\begin{abstract}
Stack-augmented recurrent neural networks (RNNs)  have been of interest to the deep learning community for some time.
However, the difficulty of training memory models remains a problem obstructing the widespread use of such models.
In this paper, we propose the Ordered Memory architecture. 
Inspired by Ordered Neurons \citep{shen2018ordered}, we introduce a new attention-based mechanism and use its cumulative probability to control the writing and erasing operation of memory. 
We also introduce a new Gated Recursive Cell to compose lower level representations into higher level representation.
We demonstrate that our model achieves strong performance on the logical inference task \citep{bowman2015tree} and the ListOps \citep{nangia2018listops} task.
We can also interpret the model to retrieve the induced tree structure, and find that these induced structures align with the ground truth.
Finally, we evaluate our model on the Stanford Sentiment Treebank tasks \citep{socher2013recursive}, and find that it performs comparatively with the state-of-the-art methods in the literature\footnote{The code can be found at \url{https://github.com/yikangshen/Ordered-Memory}}.
\end{abstract}

\section{Introduction}

A long-sought after goal in natural language processing is to build models that account for the compositional nature of language --- granting them an ability to understand complex, unseen expressions from the meaning of simpler, known expressions \citep{montague1970universal,dowty2007compositionality}. 
Despite being successful in language generation tasks, recurrent neural networks (RNNs, \citet{elman1990finding}) fail at tasks that explicitly require and test compositional behavior \citep{lake2017generalization,loula2018rearranging}.
In particular, \citet{bowman2015tree}, and later~\citet{bahdanau2018systematic} give evidence that, by exploiting the appropriate compositional structure of the task, models can generalize better to out-of-distribution test examples.
Results from~\citet{andreas2016neural} also indicate that recursively composing smaller modules results in better representations.
The remaining challenge, however, is learning the underlying structure and the rules governing composition from the observed data alone.
This is often referred to as the \emph{grammar induction}~\citep{chen1995bayesian, cohen2011unsupervised,roark2001probabilistic,chelba2000structured,williams2018latent}.

\citet{fodor1988connectionism} claim that ``cognitive capacities always exhibit certain symmetries, so that the ability to entertain a given thought implies the ability to entertain thoughts with semantically related contents," and use the term \emph{systematicity} to describe this phenomenon.
Exploiting known symmetries in the structure of the data has been a useful technique for achieving good generalization capabilities in deep learning, particularly in the form of convolutions \citep{fukushima1980neocognitron}, which leverage parameter-sharing.
If we consider architectures used in \citet{socher2013recursive} or \citet{tai2015improved}, the same recursive operation is performed at known points along the input where the substructures are meant to be composed. 
Could symmetries in the structure of natural language data be learned and exploited by models that operate on them?



In recent years, many attempts have been made in this direction using neural network architectures~\citep{grefenstette2015learning,bowman2016fast,williams2018latent,yogatama2018memory,shen2018ordered,dyer2016recurrent}. These models typically augment a recurrent neural network with a stack and a buffer which operate in a similar way to how a shift-reduce parser builds a parse-tree.
While some assume that ground-truth trees are available for supervised learning~\citep{bowman2016fast,dyer2016recurrent}, others 
use reinforcement learning (RL) techniques to learn the optimal sequence of shift reduce actions in an unsupervised fashion~\citep{yogatama2018memory}.

To avoid some of the challenges of RL training~\citep{havrylov2019cooperative}, some approaches use a~\emph{continuous stack}~\citep{grefenstette2015learning,joulin2015inferring,yogatama2018memory}.
These can usually only perform
one push or pop action per time step, requiring different mechanisms --- akin to adaptive computation time (ACT,~\citet{graves2016adaptive,jernite2016variable}) --- to perform the right number of shift and reduce steps to express the correct parse.
In addition, continuous stack models tend to ``blur" the stack due to performing a ``soft'' shift of either the pointer to the head of the stack~\citep{grefenstette2015learning}, or all the values in the stack~\citep{joulin2015inferring,yogatama2018memory}.
Finally, while these previous models can learn to manipulate a stack, they lack the capability to~\emph{lookahead} to future tokens before performing the stack manipulation for the current time step.

In this paper, we propose a novel architecture: \emph{Ordered Memory} (OM), which includes a new memory updating mechanism and a new Gated Recursive Cell.
%
We demonstrate that our method generalizes for synthetic tasks where the ability to parse is crucial to solving them.
In the Logical inference dataset \citep{bowman2015tree}, we show that our model can systematically generalize to unseen combination of operators.
In the ListOps dataset~\citep{nangia2018listops}, we show that our model can learn to solve the task with an order of magnitude less training examples than the baselines.
The parsing experiments shows that our method can effectively recover the latent tree structure of the both tasks with very high accuracy.
We also perform experiments on the Stanford Sentiment Treebank, in both binary classification and fine-grained settings (SST-2 \& SST-5), and find that we achieve comparative results to the current benchmarks.


\section{Related Work}
Composition with recursive structures has been shown to work well for certain types of tasks.~\citet{pollack1990recursive} first suggests their use with distributed representations.
Later,~\citet{socher2013recursive} shows their effectiveness on sentiment analysis tasks.
Recent work has demonstrated that recursive composition of sentences is crucial to systematic generalisation \citep{bowman2015tree,bahdanau2018systematic}.~\citet{kuncoro2018lstms} also demonstrate that architectures like \citet{dyer2016recurrent} handle syntax-sensitive dependencies better for language-related tasks.

\citet{schutzenberger1963context} first showed an equivalence between push-down automata (stack-augmented automatons) and context-free grammars.~\citet{knuth1965translation} introduced the notion of a shift-reduce parser that uses a stack for a subset of formal languages that can be parsed from left to right. This technique for parsing has been applied to natural language as well:~\citet{shieber1983sentence} applies it to English, using assumptions about how native English speakers parse sentences to remove ambiguous parse candidates. More recently, \citet{maillard2017jointly} shows that a soft tree could emerge from all possible tree structures through back propagation.

The idea of using neural networks to control a stack is not new.~\citet{zeng1994discrete} uses gradient estimates to learn to manipulate a stack using a neural network.~\citet{das1992learning} and~\citet{mozer1993connectionist} introduced the notion of a \emph{continuous stack} in order for the model to be fully differentiable.
Much of the recent work with stack-augmented networks built upon the development of neural attention \citep{graves2013generating,bahdanau2014neural,weston2014memory}.
\citet{graves2014neural} proposed methods for reading and writing using a head, along with a ``soft" shift mechanism. Apart from using attention mechanisms, \citet{grefenstette2015learning} proposed a neural stack where the push and pop operations are made to be differentiable, which worked well in synthetic datasets. \citet{yogatama2016learning} proposes RL-SPINN where the discrete stack operations are directly learned by reinforcement learning.




\section{Model}

\begin{figure}
    \centering
        \includegraphics[width=1\textwidth]{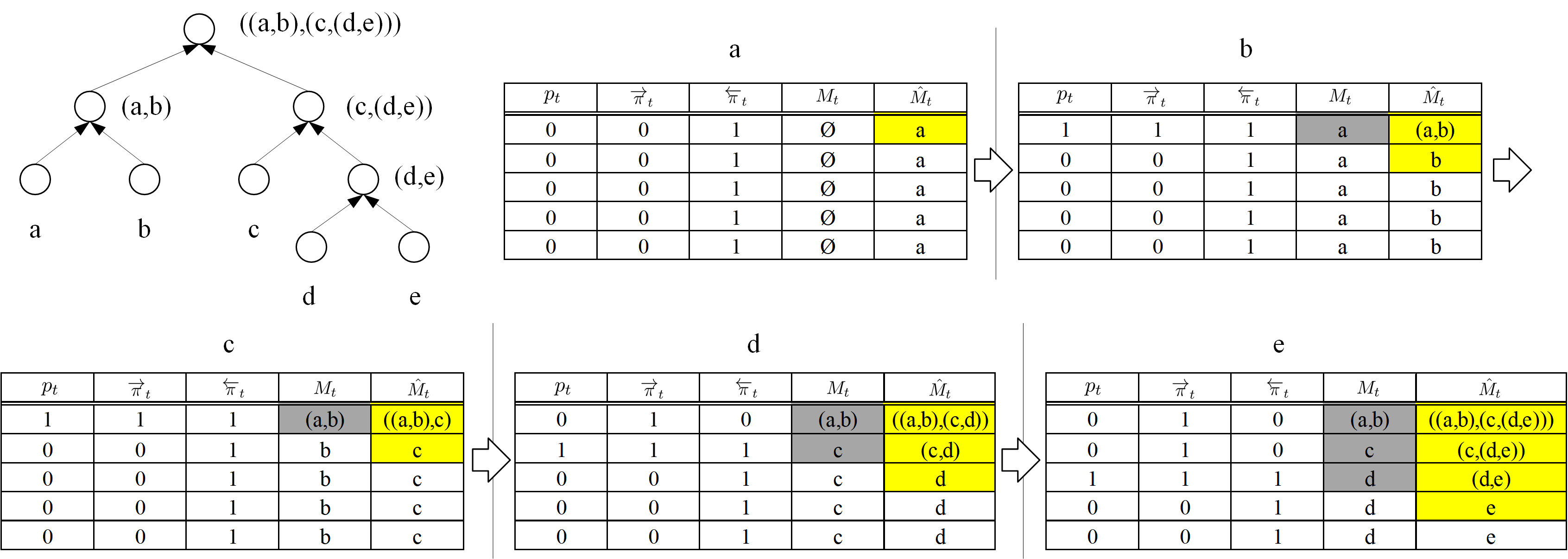}
    
    \caption{
    An example run of the OM model. 
    Let the input sequence $a,b,c,d,e$ and its hierarchical structure be as shown in the figure.
    Ideally, the OM model will output the values shown in the above tables.
    The occupied slots in $M_t$ are highlighted in gray.
    The yellow slots in $\hat{M}_t$ are slots that can be attended on in time-step $t+1$.
    At the first time-step ($t=1$), the model will initialize the candidate memory $\hat{M}_1$ with input $a$ and the memory $M_0$ with zero vectors.
    At $t=2$, the model attends on the last memory slot to compute $M_1$ (Eqn. \ref{eq:write_m}), followed by $\hat{M}_2$ (Eqn. \ref{eq:write_mhat}).
    At $t=3$, given the input $c$, the model will attend on the last slot.
    Consequently the memory slot for $b$ is erased by $\cp_3$.
    Given Eqns. \ref{eq:cell} and \ref{eq:write_mhat}, our model will recursively compute every slot in the candidate memory $\hat{M}_t^i$ to include information from $\hat{M}_t^{i-1}$ and $M_{t-1}^i$.
    Since the $\cell(\cdot)$ function only takes 2 inputs, the actual computation graph is a binary tree.
    }
    \label{fig:model}
\end{figure}


The OM model actively maintains a stack and processes the input from left to right, with a one-step lookahead in the sequence.
This allows the OM model to decide the local structure more accurately, much like a shift-reduce parser ~\citep{knuth1965translation}.

At a given point $t$ in the input sequence $\boldsymbol{x}$ (the $t$-th time-step), we have a memory of candidate sub-trees spanning the non-overlapping sub-sequences in $x_1,\hdots, x_{t-1}$, with each sub-tree being represented by one slot in the memory stack.
We also maintain a memory stack of sub-trees that contains $x_1,\hdots, x_{t-2}$.
We use the input $x_t$ to choose its parent node from our previous candidate sub-trees.
The descendant sub-trees of this new sub-tree (if they exist) are removed from the memory stack, and this new sub-tree is then added.
We then build the new candidate sub-trees that include $x_t$ using the current input and the memory stack.
In what follows, we describe the OM model in detail.
To facilitate a clearer description, a discrete attention scheme is assumed, but only ``soft" attention is used in both the training and evaluation of this model.

Let $D$ be the dimension of each memory slot and $N$ be the number of memory slots. At time-step $t$, the model takes four inputs:
\begin{itemize}
    \item $M_{t-1}$: a memory matrix of dimension $N \times D$, where each occupied slot is a distributed representation for sub-trees spanning the non-overlapping subsequences in $x_1, ..., x_{t-2}$;
    \item $\hat{M}_{t-1}$: a matrix of dimension $N \times D$ that contains representations for candidate subtrees that include the leaf node $x_{t-1}$;
    \item $\  \cp_{t-1}$: a vector of dimension $N$, where each element indicate whether the respective slot in $M_{t-1}$ occupied by a subtree.
    \item $x_t$: a vector of dimension $D_{in}$, the input at time-step $t$.
\end{itemize}


The model first transforms $x_t$ to a $D$ dimension vector.
\begin{equation}
    \tilde{x}_t = LN(W x_t + b) \label{eq:input_proj}
\end{equation}
where $LN(\cdot)$ is the layer normalization function \citep{ba2016layer}.

To select the candidate representations from $\hat{M}_{t-1}$, the model uses $\tilde{x}_t$ as its query to attend on $\hat{M}_{t-1}$:
\begin{align}
    p_t &= \mathrm{Att}(\tilde{x}_t, \hat{M}_{t-1}, \  \cp_{t-1}) \\
    \cp^{i}_t &= \sum_{j\leq i}p^{j}_t \\
    \rcp^{i}_t &=  \sum_{j\geq i}p^{j}_t \label{eq:rcp}
\end{align}
where $\mathrm{Att}(\cdot)$ is a masked attention function, $\cp_{t-1}$ is the mask, $p_t$ is a distribution over different memory slots in $\hat{M}_{t-1}$, and $p_t^j$ is the probability on the $j$-th slot.
The attention mechanism will be described in section \ref{sec:stickbreaking}.
Intuitively, $p_t$ can be viewed as a pointer to the head of the stack, $\cp_t$ is an indicator value over where the stack exists, and $\rcp_t$ is an indicator over where the top of the stack is and where it is non-existent. 

To compute $M_{t}$, we combine $\hat{M}_{t-1}$ and $M_{t-1}$ through:
\begin{eqnarray}
    M_{t}^{i} = M_{t-1}^{i} \cdot (1 - \rcp)^{i} + \hat{M}_{t-1}^{i} \cdot \rcp_t^{i}, \quad \forall i \label{eq:write_m}
\end{eqnarray}

\begin{wrapfigure}{r}{0.5\textwidth}
    \begin{minipage}{0.5\textwidth}
    \begin{algorithm}[H]
        \SetAlgoLined
        \KwData{$x_1, ..., x_T$}
        \KwResult{$o_T^N$}
     
        initialize $M_{0}, \hat{M}_0$\;
        \For{$i\leftarrow 1$ \KwTo $T$}{
            $\tilde{x}_t = LN(W x_t + b)$\;
            $p_t = \mathrm{Att}(\tilde{x}_t, \hat{M}_{t-1},   \cp_{t-1})$\;
            $\cp^{i}_t = \sum_{j\leq i}p^{j}_t$\;
            $\rcp^{i}_t =  \sum_{j\geq i}p^{j}_t$\;
            $\hat{M}^0_t = \tilde{x}_t$\;
            \For{$i\leftarrow 1$ \KwTo $N$}{
                $M_{t}^{i} = M_{t-1}^{i} \cdot (1 - \rcp_t)^{i} +
                \hat{M}_{t-1}^{i} \cdot \rcp_t^{i}$\;
                $o_t^i = \cell(M_{t}^i, \hat{M}_{t}^{i-1})$\;
                $\hat{M}_t^i = \tilde{x}_t \cdot (1-  \cp_t)^{i} + o_t^i \cdot   \cp_t^{i}$\;
            }
        }
        \Return{$o_T^N$}\;
        \caption{
        Ordered Memory algorithm. The attention function $\mathrm{Att}(\cdot)$ is defined in section \ref{sec:stickbreaking}. The recursive cell function $\mathrm{cell}(\cdot)$ is defined in section \ref{sec:cell}.}
        \label{algo:OM}
    \end{algorithm}
    \end{minipage}
\end{wrapfigure}

Suppose $p_t$ points at a memory slot $y_t$ in $\hat{m}$. 
Then $\rcp_t$ will write $\hat{M}_{t-1}^i$ to $M_{t}^i$ for $i \leq y_t$, and $(1 - \rcp_t)$ will write $M_{t-1}^i$ to $M_{t}^i$ for $i > y_t$.
In other words, Eqn. \ref{eq:write_m} copies everything from $M_{t-1}$ to the current timestep, up to the where $p_t$ is pointing.

We believe that this is a crucial point that differentiates our model from past stack-augmented models like \citet{yogatama2016learning} and \citet{joulin2015inferring}.
Both constructions have the 0-th slot as the top of the stack, and perform a convex combination of each slot in the memory / stack given the action performed.
More concretely, a distribution over the actions that is not sharp (e.g. 0.5 for pop) will result in a weighted sum of an un-popped stack and a pop stack, resulting in a blurred memory state.
Compounded, this effect can make such models hard to train.
In our case, because $(1 - \rcp_t)^{i}$ is non-decreasing with $i$, its value will accumulate to 1 at or before $N$.
This results in a full copy, guaranteeing that the earlier states are retained.
This full retention of earlier states may play a part in the training process, as it is a strategy also used in \citet{gulcehre2017memory}, where all the memory slots are filled before any erasing or writing takes place.

To compute candidate memories for time step $t$, we recurrently update all memory slots with
\begin{align}
    o^i &= \cell(M_{t}^i, \hat{M}_{t}^{i-1}) \label{eq:cell} \\
    \hat{M}_t^i &= \tilde{x}_t \cdot (1 - \cp_t)^{i+1} + o^i_t \cdot \cp_t^{i},  \forall i
    \label{eq:write_mhat}
\end{align}
where $\hat{M}^0_t$ is $x_t$.
The $\cell(\cdot)$ function can be seen as a recursive composition function in a recursive neural network \citep{socher2013recursive}. We propose a new cell function in section \ref{sec:cell}. 

The output of time step $t$ is the last memory slot $\hat{M}_t^N$ of the new candidate memory, which summarizes all the information from $x_1, ..., x_t$ using the induced structure.
The pseudo-code for the OM algorithm is shown in Algorithm \ref{algo:OM}.

\subsection{Masked Attention} \label{sec:stickbreaking}

Given the projected input $\tilde{x}_t$ and candidate memory $\hat{M}_{t-1}^{i}$.
We feed every $( \tilde{x}_t, \hat{M}_{t-1}^{i} )$ pair into a feed-forward network:
\begin{eqnarray}
    \alpha_t^i &=& \frac{ \mathbf{w}^{Att}_2 ~ \mathrm{tanh} \left( \mathbf{W}^{Att}_1  
    \left[ 
    \begin{matrix}
        \hat{M}_{t-1}^i \\ 
        \tilde{x}_t
    \end{matrix} 
    \right] 
    + b_1 \right)+ b_2 }{\sqrt{N}} \\
    \beta_t^i &=& \exp \left( \alpha_t^i - \max_j \alpha_t^j \right)
\end{eqnarray}
where $\mathbf{W}^{Att}_1$ is $N \times 2N$ matrix, $\mathbf{w}^{Att}_2$ is $N$ dimension vector, and the output $\beta^i_t$ is a scalar. 
The purpose of dividing by $\sqrt{N}$ is to scale down the logits before softmax is applied, a technique similar to the one seen in \citet{vaswani2017attention}.
We further mask the $\beta_t$ with the cumulative probability from the previous time step to prevent the model attending on non-existent parts of the stack:
\begin{eqnarray}
    \hat{\beta}_t^i = \beta_t^i   \cp_{t-1}^{i+1} \label{eq:cum_prod}
\end{eqnarray}
where $  \cp_{t-1}^{N+1} = 1$ and $  \cp_0^{\leq N}=0$.
We can then compute the probability distribution:
\begin{eqnarray}
    p_t^i = \frac{\hat{\beta}_{t}^i}{\sum_j \hat{\beta}_{t}^j} 
    \label{eq:stickbreaking}
\end{eqnarray}
This formulation bears similarity to the method used for the multi-pop mechanism seen in \citet{yogatama2018memory}.

\subsection{Gated Recursive Cell} \label{sec:cell}
Instead of using the recursive cell proposed in TreeLSTM \citep{tai2015improved} and RNTN \citep{socher2010learning}, we propose a new gated recursive cell, which is inspired by the feed-forward layer in Transformer \citep{vaswani2017attention}. 
The inputs $M^{i}_{t}$ and $\hat{M}_t^{i-1}$ are concatenated and fed into a fully connect feed-forward network:
\begin{equation}
    \left[
    \begin{matrix}
        \vg_t^i \\ 
        \hg_t^i \\
        \cg_t^i \\
        u_t^i
    \end{matrix}
    \right]=  \mathbf{W}^{Cell}_2~ \mathrm{ReLU} \left( \mathbf{W}^{Cell}_1  \left[ 
    \begin{matrix}
        \hat{M}_t^{i-1} \\ 
        M^i_{t} 
    \end{matrix} 
    \right] 
    + b_1 \right) + b_2
\end{equation}
Like the TreeLSTM, we compute the output with a gated combination of the inputs and $u_t^i$:
\begin{eqnarray}
    o_t^i &=& LN (\sigma(\vg^i_t) \odot \hat{M}^{i-1}_t
                    + \sigma(\hg^i_t) \odot M^{i}_{t} 
                    + \sigma(\cg^i_t) \odot u^i_t ) \label{eq:gated_sum}
\end{eqnarray}
where $\vg_t^i$ is the vertical gate that controls the input from previous slot, $\hg_t^i$ is horizontal gate that controls the input from previous time step, $cg_t^i$ is cell gate that control the $u^i_t$, $o^i_t$ is the output of cell function, and $LN(\cdot)$ share the same parameters with the one used in the Eqn. \ref{eq:input_proj}.

\subsection{Relations to ON-LSTM and Shift-reduce Parser}\label{sec:relationship}
Ordered Memory is implemented following the principles introduced in Ordered Neurons \citep{shen2018ordered}.
Our model is related to ON-LSTM in several aspects:
1) The memory slots are similar to the chunks in ON-LSTM, when a higher ranking memory slot is forgotten/updated, all lower ranking memory slots should likewise be forgotten/updated;
2) ON-LSTM uses the monotonically non-decreasing master forget gate to preserve long-term information while erasing short term information, the OM model uses the cumulative probability $\cp_t$;
3) Similarly, the master input gate used by ON-LSTM to control the writing of new information into the memory is replaced with the reversed cumulative probability $\rcp_t$ in the OM model.

At the same time, the internal mechanism of OM can be seen as a continuous version of a shift-reduce parser. 
At time step $t$, a shift-reduce parser could perform zero or several reduce steps to combine the heads of stack, then shift the word $t$ into stack.
The OM implement the reduce step with Gated Recursive Cell. 
It combines $\hat{M}_t^{i-1}$, the output of previous reduce step, and $M_t^i$, the next element in stack, into $\hat{M}_t^{i}$, the representation for new sub-tree.
The number of reduce steps is modeled with the attention mechanism. 
The probability distribution $p_t$ models the position of the head of stack after all necessary reduce operations are performed.
The shift operations is implemented as copying the current input word $x_t$ into memory.

The upshot of drawing connections between our model and the shift-reduce parser is interpretability: We can approximately infer the computation graph constructed by our model with Algorithm \ref{algo:parsing} (see appendix).
The algorithm can be used for the latent tree induction tasks in \citep{williams2018latent}.

\section{Experiments}
We evaluate the tree learning capabilities of our model on two datasets: logical inference~\citep{bowman2015tree} and ListOps~\citep{nangia2018listops}.
In these experiments, we infer the trees with our model by using Alg.~\ref{algo:parsing} and compare them with the ground-truth trees used to generate the data.
We evaluate parsing performance using the $F_1$ score\footnote{All parsing scores are given by \texttt{Evalb}~\url{https://nlp.cs.nyu.edu/evalb/}}. We also evaluate our model on Stanford Sentiment Treebank (SST), which is the sequential labeling task described in ~\citet{socher2013recursive}.

\subsection{Logical Inference}

\begin{table}
\small
    \centering
    \setlength{\tabcolsep}{3.5pt} 
    \caption{
    Test accuracy of the models, trained on operation lengths of $\leq 6$, with their out-of-distribution results shown here (lengths 7-12).
    We ran 5 different runs of our models, giving the error bounds in the last row.
    The $F_1$ score is the parsing score with respect to the ground truth tree structure. 
    The TreeCell is a recursive neural network based on the Gated Recursive Cell function proposed in section \ref{sec:cell}.
    For the Transformer and Universal Transformer, we follow the entailment architecture introduced in \citet{radford2018improving}. 
    The model takes \texttt{<start> sentence1 <delim> sentence2 <extract>} as input, then use the vector representation for \texttt{<extract>} position at last layer for classification.
    $^*$The results for RRNet were taken from \citet{jacob2018learning}.
    }
    \begin{tabular}{l c cccccc  c ccc}
    \toprule
    \textbf{Model} &&  \multicolumn{6}{c}{\textbf{Number of Operations}} 
          &&  \multicolumn{3}{c}{\textbf{Sys. Gen.}} \\
          
          && 7 & 8 & 9 & 10 & 11 & 12
          && A & B & C \\
    \midrule
    \multicolumn{12}{l}{\emph{Sequential sentence representation}}\\
    LSTM     && 88 & 84 & 80 & 78 & 71 & 69 && 84 & 60 & 59 \\
    RRNet*    && 84 & 81 & 78 & 74 & 72 & 71 && -- & -- & -- \\
    ON-LSTM  && 91 & 87 & 85 & 81 & 78 & 75 && 70 &	63 & 60 \\
    \cmidrule{1-12}
    \multicolumn{12}{l}{\emph{Inter sentence attention}}\\
    Transformer && 51 & 52 & 51 & 51 & 51 & 48 && 53	 & 51	& 51\\
    Universal Transformer && 51 & 52 & 51 & 51 & 51 & 48 && 53	 & 51	& 51 \\
    \cmidrule{1-12}
    \multicolumn{12}{l}{\emph{Our model}}\\
    Accuracy && 98 $\pm$ 0.0 & 97 $\pm$ 0.4 & 96 $\pm$ 0.5 & 94 $\pm$ 0.8 & 93 $\pm$ 0.5 & 92 $\pm$ 1.1
             && 94 &	91 &	81\\
    Parsing $F_1$ && \multicolumn{6}{c}{$84.3 \pm 14.4$}\\
    \cmidrule(lr){1-12}
    \multicolumn{12}{l}{\emph{Ablation tests}}\\
    \st{$\mathrm{cell}(\cdot)$} TreeRNN Op. & & 69 & 67 & 65 & 61 & 57 & 53
    && -- & -- & -- \\
    \cmidrule{1-12}
    \multicolumn{12}{l}{\emph{Recursive NN + ground-truth structure}}\\
    TreeLSTM && 94 & 92 & 92 & 88 & 87 & 86 && 91 & 84 & 76 \\
    TreeCell && 98 & 96 & 96 & 95 & 93 & 92 && 95 & 95 & 90 \\
    TreeRNN  && 98 & 98 & 97 & 96 & 95 & 96 && 94 &	92 & 86 \\
    \bottomrule
    \end{tabular}
    \label{tab:proplogparse}
\end{table}



The logical inference task described in~\citet{bowman2015tree} has a vocabulary of six words and three logical operations, \texttt{or}, \texttt{and}, \texttt{not}. 
The task is to classify the relationship of two logical clauses into seven mutually exclusive categories.
We use a multi-layer perceptron (MLP) with $(h_1, h_2, h_1 \circ h_2, |h_1 - h_2|)$ as input, where $h_1$ and $h_2$ are the $\hat{M}^N_T$ of their respective input sequences.
We compare our model with LSTM, RRNet \citep{jacob2018learning}, ON-LSTM \citep{shen2018ordered}, Tranformer \citep{vaswani2017attention}, Universal Transformer \citep{dehghani2018universal}, TreeLSTM \citep{tai2015improved}, TreeRNN \citep{bowman2015tree}, and TreeCell.
We used the same hidden state size for our model and baselines for proper comparison. 
Hyper-parameters can be found in Appendix \ref{appendix:hyperparameters}.
The model is trained on sequences containing up to $6$ operations and tested on sequences with higher number (7-12) of operations.

The Transformer models were implemented by modifying the code from the Annotated Transformer\footnote{\url{http://nlp.seas.harvard.edu/2018/04/03/attention.html}}.
The number of Transformer layers are the same as the number of slots in Ordered Memory.
Unfortunately, we were not able to successfully train a Transformer model on the task, resulting in a model that only learns the marginal over the labels.
We also tried to used Transformer as a sentence embedding model, but to no avail.
\citet{tran2018importance} achieves similar results, suggesting this could be a problem intrinsic to self-attention mechanisms for this task.

\paragraph{Length Generalization Tests}
The TreeRNN model represents the best results achievable if the structure of the tree is known.
The TreeCell experiment was performed as a control to isolate the performance of using the $\cell(\cdot)$ composition function versus using both using $\cell(\cdot)$ and learning the composition with OM.
The performance of our model degrades only marginally with increasing number of operations in the test set, suggesting generalization on these longer sequences never seen during training.

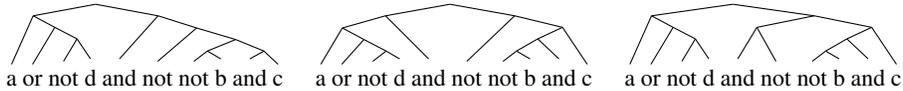
\begin{figure}[t]
\small
\begin{center}
    \begin{forest}
		shape=coordinate,
		where n children=0{
			tier=word
		}{},
		nice empty nodes
[[[a], [[or], [[not], [d]]]], [[and], [[not], [[[not], [b]], [[and], [c]]]]]]
	\end{forest}
	\quad
	\begin{forest}
		shape=coordinate,
		where n children=0{
			tier=word
		}{},
		nice empty nodes
[[[[a], [[or], [[not], [d]]]], [and]], [[not], [[[[not], [b]], [and]], [c]]]]
	\end{forest}
	\quad 
	\begin{forest}
		shape=coordinate,
		where n children=0{
			tier=word
		}{},
		nice empty nodes
 [[[a], [[or], [[not], [d]]]], [[[and], [not]], [[[[not], [b]], [and]], [c]]]]
	\end{forest}
\end{center}

\caption{
Variations in induced parse trees under different runs of the logical inference experiment. 
The left most tree is the ground truth and one of induced structures.
We have removed the parentheses in the original sequence for this visualization.
It is interesting to note that the different structures induced by our model are all valid computation graphs to produce the correct results.
}
\label{fig:trees}
\end{figure}

\paragraph{Parsing results}
There is a variability in parsing performance over several runs under different random seeds, but the model's ability to generalize to longer sequences remains fairly constant.
The model learns a slightly different method of composition for consecutive operations.
Perhaps predictably, these are variations that do not affect the logical composition of the subtrees.
The source of different parsing results can be seen in Figure \ref{fig:trees}.
The results suggest that these latent structures are still valid computation graphs for the task, in spite of the variations.

\begin{table*}[t]
\small
    \caption{
    Partitions of the Logical Inference task from \citet{bowman2014recursive}. Each partitions include a training set filtered out all data points that match the rule indicated in \textbf{Excluded}, and a test set formed by matched data points.
    }
    \centering
    \begin{tabular}{r l r l}
    \toprule
    \textbf{Part.} & \textbf{Excluded} & 
    \textbf{Training set size}  &
    \textbf{Test set example} \\
    \midrule
   
    A & \texttt{* ( and (not a) ) *} & 
        128,969 &
        \texttt{f (and (not a))} \\
    B & \texttt{* ( and (not *) ) *} &
        87,948  &
        \texttt{c (and (not (a (or b))))}  \\
    C & \texttt{* ( \{and,or\} (not *) ) *} &
          51,896 &
        \texttt{a (or (e (and c)))}\\
    \midrule
    Full &  & 135,529 &  \\
    \bottomrule
    \end{tabular}

    \label{tab:ablation_tests}
\end{table*}

\paragraph{Systematic Generalization Tests}
Inspired by \citet{loula2018rearranging}, we created three splits of the original logical inference dataset with increasing levels of difficulty.
Each consecutive split removes a superset of the previously excluded clauses, creating a harder generalization task.
Each model is then trained on the ablated training set, and tested on examples unseen in the training data.
As a result, the different test splits have different numbers of data points.
Table \ref{tab:ablation_tests} contains the details of the individual partitions.

The results are shown in the right section of Table \ref{tab:proplogparse} under Sys. Gen.
Each column labeled A, B, and C are the model's aggregated accuracies over the unseen operation lengths.
As with the length generalization tests, the models with the known tree structure performs the best on unseen structures, while sequential models degrade quickly as the tests get harder.
Our model greatly outperforms all the other sequential models, performing slightly below the results of TreeRNN and TreeCell on the different partitions.

Combined with the parsing results, and our model's performance on these generalization tests, we believe this is strong evidence that the model has both (i) learned to exploit symmetries in the structure of the data by learning a good $\cell(\cdot)$ function, and (ii) learned where and how to apply said function by operating its stack memory.

\subsection{ListOps}
\citet{nangia2018listops} build a dataset with nested summary operations on lists of single digit integers.
The sequences comprise of the operators \texttt{MAX}, \texttt{MIN}, \texttt{MED}, and \texttt{SUM\_MOD}.
The output is also an integer in $[0, 9]$
As an example, the input: \texttt{[MAX 2 9 [MIN 4 7 ] 0 ]} has the solution \texttt{9}.
As the task is formulated to be easily solved with a correct parsing strategy, the task provides an excellent test-bed to diagnose models that perform tree induction.
The authors binarize the structure by choosing the subtree corresponding to each list to be left-branching: the model would first take into account the operator, and then proceed to compute the summary statistic within the list. A right-branching parse would require the entire list to be maintained in the model's hidden state. 

\begin{figure}
    \centering
    \begin{subfigure}[b]{0.45\textwidth}
        \small
        \begin{tabular}{l cc }
            \toprule
            \textbf{Model} & \textbf{Accuracy}  & $F_1$\\
            \midrule
            \multicolumn{3}{l}{\emph{Baselines}} \\
            LSTM*                   & 71.5\(\pm\)1.5  & --\\
            RL-SPINN*               & 60.7\(\pm\)2.6  & 71.1 \\
            Gumbel Tree-LSTM*       & 57.6\(\pm\)2.9  & 57.3 \\
            Transformer             & 57.4\(\pm\)0.4  & -- \\
            Universal Transformer   & 71.5\(\pm\)7.8  & -- \\
            \citet{havrylov2019cooperative}  & 99.2\(\pm\)0.5 & --\\
            \midrule
            Ours & \bf 99.97\(\pm\)0.014 & \bf 100 \\
            \cmidrule(lr){1-3} 
            \multicolumn{3}{l}{\emph{Ablation tests}} \\
            \st{$\mathrm{cell}(\cdot)$} TreeRNN Op.  & 63.1 & -- \\
            \bottomrule
        \end{tabular}
        \caption{}
        \label{table:listops comparison}
    \end{subfigure}
    \quad
    \begin{subfigure}[b]{0.45\textwidth}
        \includegraphics[width=1\textwidth]{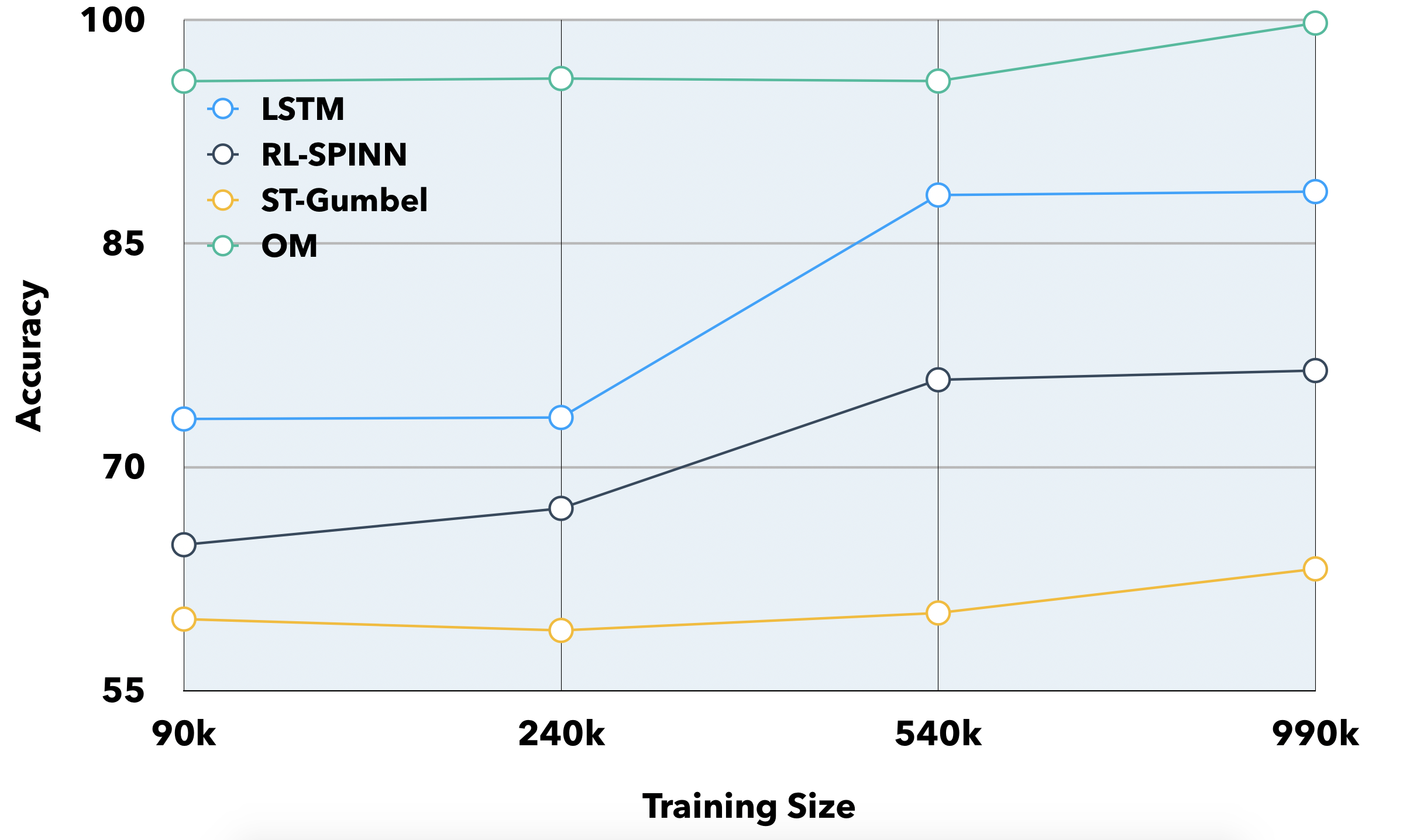}
        \caption{}
        \label{fig:listops_gen}
    \end{subfigure}
    \caption{
    (a) shows the accuracy of different models on the ListOps dataset. All models have $128$ dimensions. Results for models with * are taken from \citet{nangia2018listops}.
    (b) shows our model accuracy on the ListOps task when varying the the size of the training set.
    }
\end{figure}

Our model achieves 99.9\% accuracy, and an $F_1$ score of 100\% on the model's induced parse tree~(See Table \ref{table:listops comparison}).
This result is consistent across 3 different runs of the same experiment. In \citet{nangia2018listops}, the authors perform an experiment to verify the effect of training set size on the latent tree models.
As the latent tree models (RL-SPINN and ST-Gumbel) need to parse the input successfully to perform well on the task, the better performance of the LSTM than those models indicate that the size of the dataset does not affect the ability to learn to parse much for those models.
Our model seems to be more data efficient and solves the task even when only training on a subset of 90k examples (Fig.~\ref{fig:listops_gen}).

\subsection{Ablation studies}
We replaced the $\mathrm{cell}(\cdot)$ operator with the RNN operator found in TreeRNN, which is the best performing model that explicitly uses the structure of the logical clause.
In this test, we find that the TreeRNN operator results in a large drop across the different tasks.
The detailed results for the ablation tests on both the logical inference and the ListOps tasks are found in Table \ref{tab:proplogparse} and \ref{table:listops comparison}.

\subsection{Stanford Sentiment Treebank}

The Stanford Sentiment Treebank is a classification task described in \citet{socher2013recursive}.
There are two settings: SST-2, which reduces the task down to a positive or negative label for each sentence (the neutral sentiment sentences are ignored), and SST-5, which is a fine-grained classification task which has 5 labels for each sentence.

\begin{table*}[t]
\caption{Accuracy results of models on the SST.}\label{table:sst}
\centering
\begin{tabular}{l c c}
\toprule
    &  SST-2 & SST-5 \\
\midrule
  \multicolumn{3}{l}{\emph{Sequential sentence representation \& other methods}}\\
\citet{radford2017learning} & 91.8 & 52.9 \\
\citet{peters2018deep} & -- & 54.7\\
\citet{Brahma2018ImprovedSM} & 91.2 & 56.2 \\
\citet{devlin2018bert} & 94.9 & -- \\
\citet{liu2019multi} & 95.6 & -- \\
\midrule
  \multicolumn{3}{l}{\emph{Recursive NN + ground-truth structure}}\\
\citet{tai2015improved} & 88.0 & 51.0\\
\citet{munkhdalai2017neural} & 89.3 & 53.1\\
\citet{looks2017deep} & 89.4 & 52.3 \\
\midrule
  \multicolumn{3}{l}{\emph{Recursive NN + latent / learned structure}}\\
\citet{choi2018learning} & 90.7 & 53.7 \\
\citet{havrylov2019cooperative} & 90.2\(\pm\)0.2 & 51.5\(\pm\)0.4\\
\midrule
Ours (Glove) & 90.4 & 52.2 \\
Ours (ELMO) & 92.0 & 55.2 \\
\bottomrule
\end{tabular}
\end{table*}

Current state-of-the-art models use pretrained contextual embeddings \cite{radford2018improving,mccann2017learned,peters2018deep}.
Building on ELMO \cite{peters2018deep}, we achieve a performance comparable with the current state-of-the-art for both SST-2 and SST-5 settings.
However, it should be noted that our model is a sentence representation  model.
Table \ref{table:sst} lists our and related work's respective performance on the SST task in both settings.

\section{Conclusion}
In this paper, we introduce the Ordered Memory architecture.
The model is conceptually close to previous stack-augmented RNNs, but with two important differences: 1) we replace the pop and push operations with a new writing and erasing mechanism inspired by Ordered Neurons \citep{shen2018ordered};
2) we also introduce a new Gated Recursive Cell to compose lower level representations into higher level one.
On the logical inference and ListOps tasks, we show that the model learns the proper tree structures required to solve them.
As a result, the model can effectively generalize to longer sequence and combination of operators that is unseen in the training set, and the model is data efficient.
We also demonstrate that our results on the SST are comparable with state-of-the-art models.

\bibliographystyle{plainnat}
\bibliography{reference}

\newpage
\appendix
\section{Tree induction algorithm}
\begin{algorithm}[h]
    \SetAlgoLined
    \KwData{$p_1, ..., p_T$}
    \KwResult{$\mathbf{T}$}
    initialize $\mathrm{queue}=[w_2,...,w_T]$ \\ 
    $\quad \mathrm{stack}=[w_1], h=\mathrm{argmax}(p_1)-1$\;
    \For{$i\leftarrow 2$ \KwTo $T$}{
        $y_i=\mathrm{argmax}(p_i)$\;
        $d = y_i - h$\;
        \If{$d > 0$}{
            \For{$j\leftarrow 1$ \KwTo $d$}{
                \If{$\mathrm{len}(\mathrm{stack}) < 2$}{
                    \textbf{Break}\;
                }
                $e_1=\mathrm{stack}.\mathrm{pop}()$\;
                $e_2=\mathrm{stack}.\mathrm{pop}()$\;
                $\mathrm{stack}.\mathrm{push}(\mathrm{node}(e_1,e_2))$\;
            }
        }
        $\mathrm{stack}.\mathrm{push}(\mathrm{queue}.\mathrm{popleft}())$\;
        $h=y_i-1$\;
    }
    \While{$\mathrm{len(stack)}>2$}{
        $e_1=\mathrm{stack}.\mathrm{pop}()$\;
        $e_2=\mathrm{stack}.\mathrm{pop}()$\;
        $\mathrm{stack}.\mathrm{push}(\mathrm{node}(e_1,e_2))$\;
    }
    \caption{Shift-reduce parsing algorithm for generate parsing tree from Ordered Memory model. Here we greedily choose the $\mathrm{argmax}(p_t)$ as the head of stack for each slot.}
    \label{algo:parsing}
\end{algorithm}

\section{Hyperparameters} \label{appendix:hyperparameters}

\begin{table}[h]
    \centering
    \small
    \setlength{\tabcolsep}{3.5pt} 
    \caption{
    The hyperparameters used in the various experiments described.
    $D$ is the dimension of each slot in the memory.
    There are 4 different dropout rates for different parts of the model:
    \textit{In} dropout is applied at the embedding level input to the OM model.
    \textit{Out} dropout is applied at the layers in the MLP before the final classification task.
    \textit{Attention} dropout is applied at the layers inside stick-breaking attention mechanism.
    \textit{Hidden} dropout is applied at various other points in the OM architecture.
    }
    \begin{tabular}{l c c c c c c c c c}
    \toprule
      \textbf{Task} & \textbf{Memory size} & \textbf{\#slot} &  \multicolumn{4}{c}{\textbf{Dropout}} & \textbf{Batch size} &  \multicolumn{2}{c}{\textbf{Embedding}}
    \\
         &  &  &  \textit{In}  & \textit{Out} & \textit{Hidden} & \textit{Attention} &  &  Size & Pretrained \\
    \midrule
         Logic   & 200 & 24 & 0.1 & 0.3 & 0.2 & 0.2 & 128 & 200 & None \\
         ListOps & 128 & 21 & 0.1 & 0.2 & 0.1 & 0.3 & 128 & 128 & None \\
         SST(Glove)     & 300 & 15 & 0.3 & 0.4 & 0.2 & 0.2 & 128 & 300 & Glove\\
         SST(ELMO)     & 300 & 15 & 0.3 & 0.2 & 0.2 & 0.3 & 128 & 1024 & ELMo\\
    \bottomrule \\
    \end{tabular}
    
    \label{tab:hyperparameters}
\end{table}

\section{Dynamic Computation Time}
Given Eqn. \ref{eq:write_mhat}, we can see that some $o^i_t$s are multiplied with $\cp_t^i$.
It may not necessary to compute the cell function (Eqn. \ref{eq:cell}) if the cumulative probability $  \cp_t^i$ is smaller than a certain threshold.
This threshold actively controls the number of computation steps that we need to perform for each time step.
In our experiments, we set the threshold to be $10^{-5}$.
This idea of dynamically modulating the number of computational step is similar to Adaptive Computation Time (ACT) in \citet{graves2016adaptive}, which attempts to learn the number of computation steps required that is dependent on the input.
However, the author does not demonstrate savings in computation time.
In \citet{tan2016towards}, the authors implement a similar mechanism, but demonstrate computational savings only at test time.

\end{document}